\pdfoutput=1

\documentclass[11pt]{article}

\usepackage[preprint]{acl}

\usepackage{times}
\usepackage{latexsym}
\usepackage[inline]{enumitem}
\usepackage[T1]{fontenc}

\usepackage[utf8]{inputenc}

\usepackage{microtype}

\usepackage{inconsolata}

\usepackage{graphicx}
\usepackage{booktabs}
\usepackage{makecell}
\usepackage{tabularx}
\usepackage{amssymb}
\usepackage{pifont}

\title{Diagnosing Visual Reasoning: Challenges, Insights, and a Path Forward}

\author{%
\begin{tabular}[t]{c}
    Jing Bi$^{1}$ \quad Guangyu Sun$^{2}$ \quad Ali Vosoughi$^{1}$ \quad Chen Chen$^{2}$ \quad Chenliang Xu$^{1}$
\end{tabular}\\[6pt]
$^{1}$University of Rochester \quad $^{2}$University of Central Florida\\
{\tt\small \{jing.bi, ali.vosoughi, chenliang.xu\}@rochester.edu}\\
{\tt\small guangyu@ucf.edu, chen.chen@crcv.ucf.edu} }

\begin{document}
\maketitle
\begin{abstract}
  Multimodal large language models (MLLMs) that integrate visual and textual reasoning leverage chain-of-thought (CoT) prompting to tackle complex visual tasks, yet continue to exhibit visual hallucinations and an over-reliance on textual priors. We present a systematic diagnosis of state-of-the-art vision-language models using a three-stage evaluation framework, uncovering key failure modes. To address these, we propose an agent-based architecture that combines LLM reasoning with lightweight visual modules, enabling fine-grained analysis and iterative refinement of reasoning chains. Our results highlight future visual reasoning models should focus on integrating a broader set of specialized tools for analyzing visual content. Our system achieves significant gains (+10.3 on MMMU, +6.0 on MathVista over a 7B baseline), matching or surpassing much larger models. We will release our framework and evaluation suite to facilitate future research.
\end{abstract}

\section{Introduction}
The ability to perform coherent, structured reasoning is essential for solving complex visual understanding tasks. Unlike recognition, visual reasoning requires models to integrate perceptual cues with contextual knowledge, infer relationships between entities, track logical dependencies, and arrive at conclusions that are not immediately evident from raw pixel data. This cognitive process mirrors human problem-solving, where one sequentially interprets visual inputs and iteratively verifies conclusions ~\cite{liu2025visionreasoner,zhang2025embodiedvsr,yang2025magicvqa,fu2025refocus}.

\begin{figure}[t]
  \centering
  \includegraphics[width=1\linewidth]{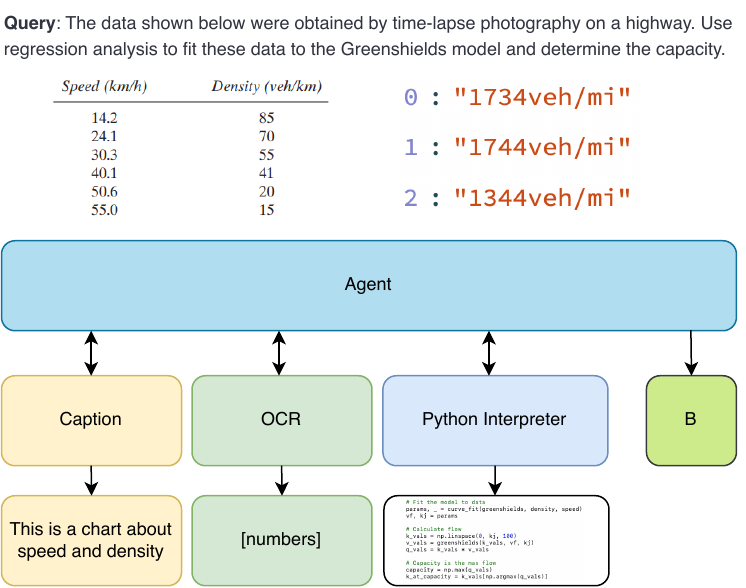}
  \caption{This image showcases our agent system that leverages a pure LLM to
  solve a visual reasoning problem using external tools. It illustrates how complex tasks, such as fitting traffic speed-density data to the Greenshields model, can offload substantial token usage to a code interpreter,
  highlighting an efficient division of labor between perception, reasoning and computation.}
  \label{fig:diagnostic-framework}
  \vspace{-5mm}
\end{figure}
\begin{table}[t]
\centering
\small
\setlength{\tabcolsep}{2pt}
\renewcommand{\arraystretch}{1.05}
\caption{Comparison of our diagnostic agent with prior modular systems: MM-ReAct~~\cite{yang2023mmreact}, MC-tree~~\cite{yao2024mulberry}.}
\label{tab:delta-table}
\resizebox{0.97\linewidth}{!}{
\begin{tabular}{lccccc}
\textbf{System} & \makecell{Math\\OCR} & \makecell{Iterative\\Diagnosis} & \makecell{Lightweight\\Backbone} & \makecell{Python\\Interpreter} & \makecell{Backtracing\\Thought}
\\
\toprule
MM-ReAct & \ding{51} & \ding{55} & \ding{51} & \ding{55} & \ding{55} \\
MC-tree & \ding{55} & \ding{51} & \ding{51} & \ding{55} & \ding{51} \\
Ours & \ding{51} & \ding{51} & \ding{51} & \ding{51} & \ding{51} \\
\bottomrule
\end{tabular}
}
\vspace{-5mm}
\end{table}


Recent advancements in LLMs have accelerated progress in this direction with strong linguistic reasoning abilities. When extended into the multimodal domain, these capabilities enable models to interpret images, diagrams, and documents extending beyond recognition to include inference and abstraction. The emergence of Reasoning Multimodal LLMs (MLLMs), such as LLaVA-CoT~~\cite{xu2024llavacot}, LlamaV-o1~~\cite{thawakar2025llamavo1}, and Heima~~\cite{heima2025efficient}, reflects this trend and demonstrates how the fusion of vision and language models can unlock new frontiers in visual intelligence.
\begin{figure*}[!ht]
  \centering
  \includegraphics[width=0.9\linewidth]{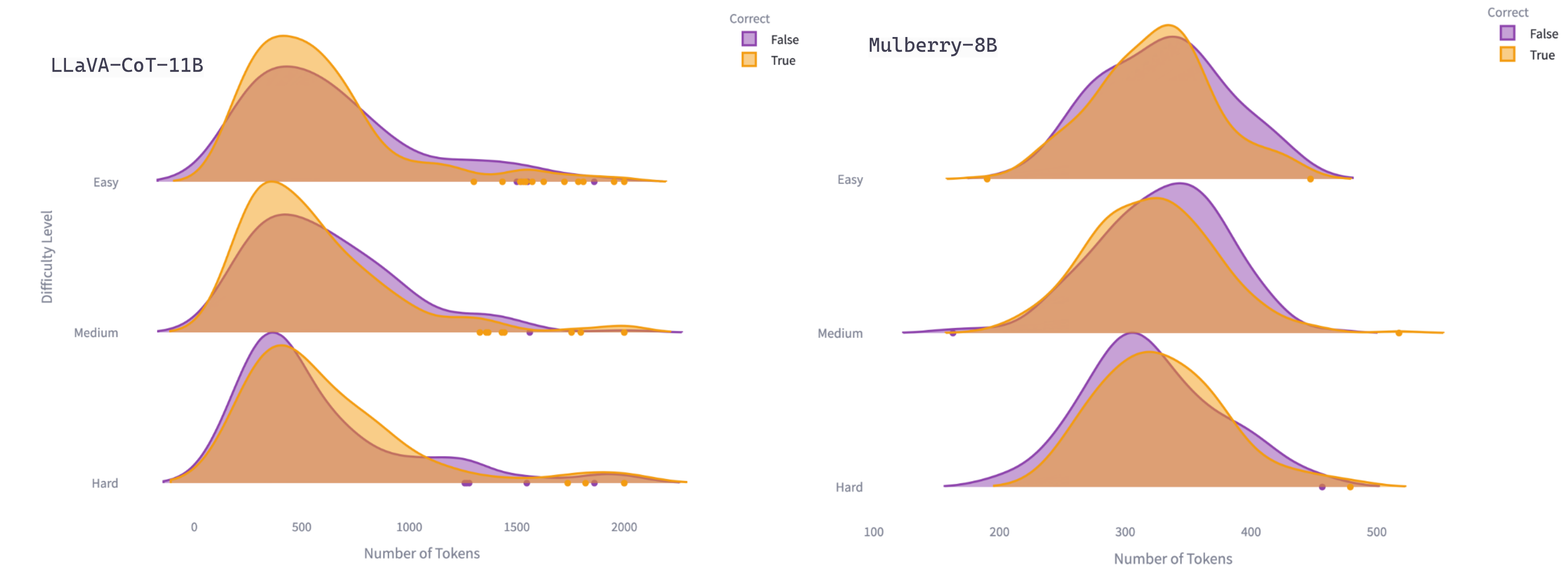}
  \caption{Comparison of token utilization and accuracy between Mulberry-7B and
  LLaVA-CoT-11B. Both models exhibit less adaptive reasoning, with token usage.
  We observe that LLaVA-CoT-11B frequently generates more tokens than Mulberry-7B, while Mulberry-7B tends to centralize its token usage around 300--350 tokens.}
  \label{fig:8vs11b-analysis}
  \vspace{-4mm}
\end{figure*}
Central to these efforts is to encourage models to produce explicit intermediate steps. This structured reasoning is particularly impactful for visual tasks, where raw perceptual data must be transformed into high-level concepts through a series of inferential stages. For example, LlamaV-o1 combines Chain-of-thought (CoT) reasoning with curriculum learning and beam search to effectively solve multi-step visual tasks, while Heima accelerates inference by encoding CoT into compact representations~\cite{shen2025efficient}

Despite these advances, models still hallucinate, producing responses not grounded in the image, and often rely too heavily on textual priors. Most models reason in a single, unidirectional pass, lacking correction or self-reflection. To address these issues, we introduce a three-stage diagnostic framework and an agent-based architecture that tightly integrates stepwise textual reasoning with lightweight visual modules, enabling fine-grained analysis of reasoning failures ~\cite{rethinking2025, progco2025, improving2025, auditing2025, attention2025}.
Unlike prior work (e.g., MC-tree~~\cite{yao2024mulberry}, MM-ReAct~~\cite{yang2023mmreact}), our agent-based framework routes tool calls at each step for explicit intervention and diagnosis, similar in spirit to recent approaches like MMCTAgent~~\cite{kumar2024mmctagent} and AgentRE~~\cite{shi2024agentre}. 
Our main contributions are:

\noindent \textbf{Diagnosis:} We present a diagnostic framework for math-centric visual reasoning, enabling granular identification and analysis of failures.\\

\vspace{-2mm}
\noindent \textbf{Agent-based Architecture:} We propose an agent-based approach that decouples perception and reasoning, integrating LLMs with visual modules for iterative reasoning, yielding substantial empirical gains over strong 7B baselines.\\

\vspace{-2mm}
\noindent \textbf{Evaluation:} We provide a comprehensive analysis of reasoning chains and release our diagnostic framework and evaluation suite to support future research in visual reasoning, enabling deeper understanding of model behaviors.

\section{Diagnostic Methodology}
Our analysis focuses on three representative visual reasoning models: QVQ (72B)~~\cite{qvq-72b-preview}, Mulberry-7B~~\cite{yao2024mulberry}(Mulberry), and LLaVA-CoT-11B~~\cite{xu2024llavacot}(LLaVA-CoT). These models span a range of parameter sizes and are selected for their popularity and relevance. We exclude OpenAI's O-series reasoning models due to the unavailability of their reasoning paths, which prevents in-depth diagnostic analysis. This selection enables a comprehensive comparison across different model scales and reasoning strategies.

\begin{figure*}[t]
  \centering
  \includegraphics[width=0.95\linewidth]{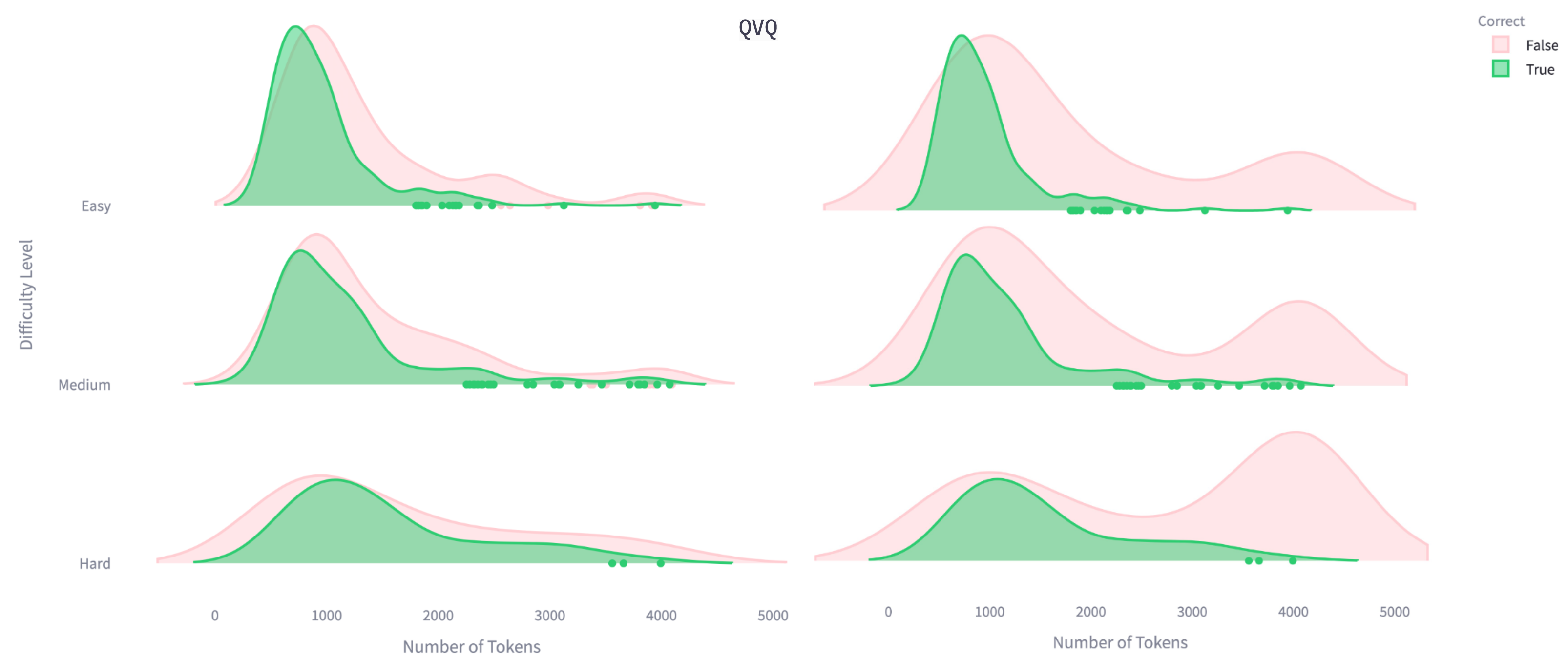}
  \caption{Token usage and accuracy trends for QVQ on MMMU. Left: Accuracy as a function of token count, showing diminishing returns and a decline beyond 2,000 tokens. Right: Distribution of token counts for correct and incorrect answers, including unfinished answers}
  \label{fig:qvq-analysis}
  \vspace{-2mm}
\end{figure*}

\begin{table}[!h]
\centering
\small
\setlength{\tabcolsep}{4pt}
\renewcommand{\arraystretch}{1.1}
\resizebox{\linewidth}{!}{
\begin{tabular}{lccc}
  \toprule
  \textbf{Dataset} & \textbf{Mulberry} & \textbf{LLaVA-CoT} & \textbf{QVQ} \\
  \midrule
  MMMU       & 52.8\%      & 55.2\%      & 60.9\% \\
  MathVista   & 63.1\%      & 57.8\%      & 65.4\% \\
  \midrule
  Base Model       & QwenVL2-7B  & Llama-3.2-7B & QwenVL2-72B \\
  \bottomrule
\end{tabular}
}
\caption{Comparison of model accuracies (\%) on MMMU and MathVista, and their base models.}
\label{tab:model-accuracy}
\vspace{-5mm}
\end{table}
\subsection{Comparative Model Analysis}

We begin our evaluation with the MMMU dataset~~\cite{yue2023mmmu} and MathVista~~\cite{lu2024mathvista}, both selected for their comprehensive problem difficulty annotations and widespread use as benchmarks for MLLMs. For the MMMU dataset, we analyze token usage across varying difficulty levels to uncover patterns in reasoning efficiency and inefficiency, as shown in Figures~\ref{fig:8vs11b-analysis},\ref{fig:qvq-analysis}.

As shown in Table~\ref{tab:model-accuracy}, Mulberry is highly succinct, with token counts clustered between 200--400, but this brevity limits its accuracy (52.8\%/63.1\%). Incorrect Mulberry responses often occur at the upper end of its token range, suggesting that rigid, template-driven reasoning can be counterproductive when the model stretches beyond its typical patterns. LLaVA-CoT, with token counts typically between 800--1,200 for correct Easy/Medium answers, achieves intermediate accuracy. On harder tasks, longer responses often correspond to incorrect answers, suggesting that concise yet sufficiently detailed reasoning chains tend to be optimal, whereas excessive verbosity may signal confusion. Our analysis shows that while more verbose reasoning can indicate higher capability, excessive token usage (beyond 2,000 for QVQ) yields diminishing returns. The best models balance detail and brevity, providing enough reasoning steps without unnecessary verbosity. QVQ’s larger size and flexible reasoning achieve the highest accuracy, but future work should aim to reduce verbosity while preserving reasoning quality.
\subsection{In-depth Examination of QVQ}
\begin{figure}[t]
  \centering
  \includegraphics[width=0.95\linewidth]{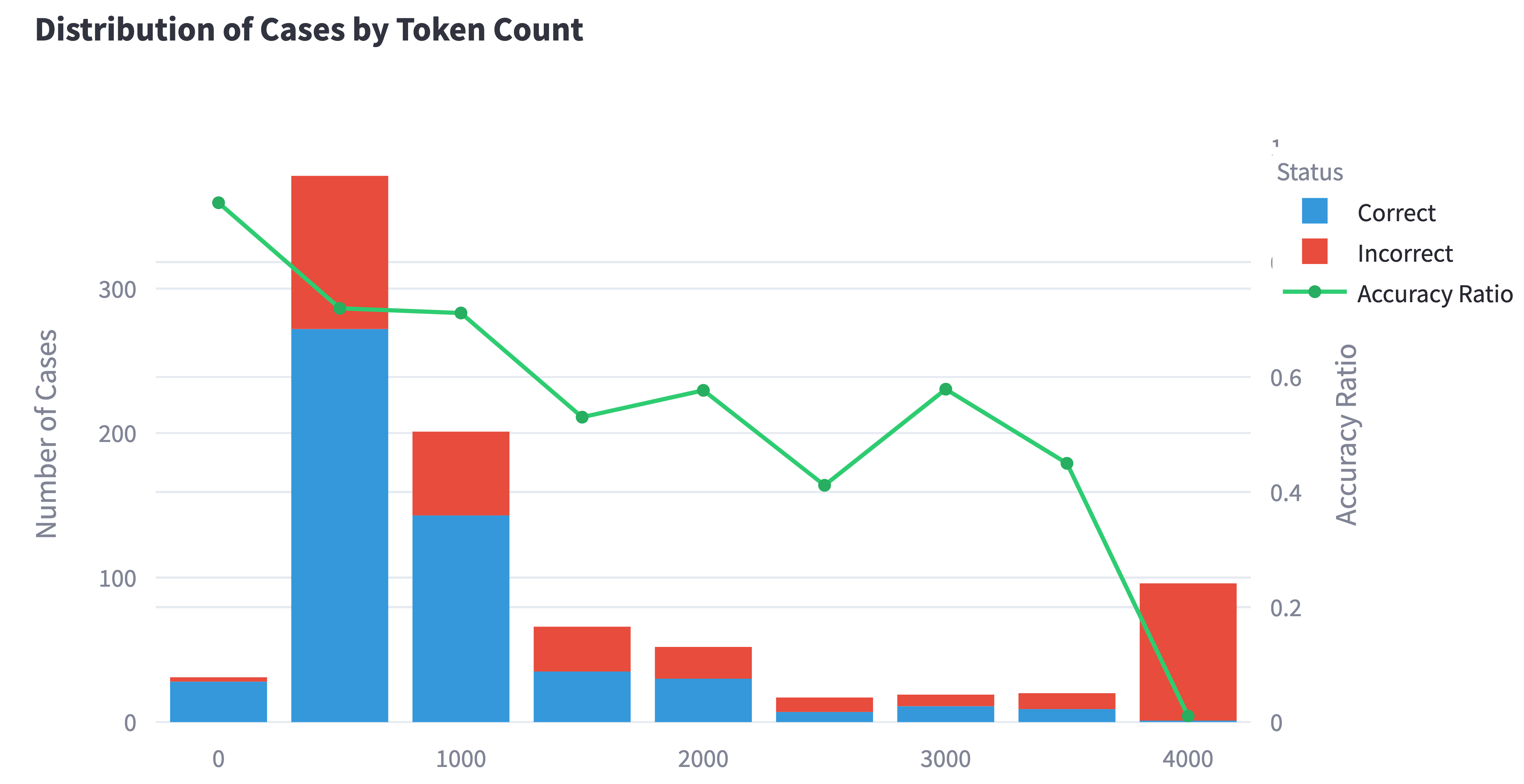}
  \caption{This chart illustrates the number of correct and incorrect cases across different token count ranges, with a green line as accuracy ratio. As token count increases, the number of cases generally decreases, and the accuracy ratio tends to decline.}
  \label{fig:diagnostic-case}
  \vspace{-2mm}
\end{figure}

As shown in Table~\ref{tab:model-accuracy}, QVQ achieves the highest accuracy—60.9\% on MMMU and 65.4\% on MathVista—outperforming both LLaVA-CoT and Mulberry. To better understand QVQ's strengths, we analyze its reasoning behavior in detail.
Figure~\ref{fig:diagnostic-case} illustrates the relationship between token count and accuracy, revealing that accuracy declines as token usage increases. This trend is partly due to our imposed hard threshold of 4,000 tokens, responses exceeding this limit are typically incomplete and considered incorrect. 
In Figure~\ref{fig:qvq-analysis}, the right panel displays the distribution of token counts for answers, while the left panel excludes unfinished responses. QVQ's superior performance comes with more tokens: it often generates 1,000--2,000 tokens for Easy/Medium tasks and over 3,000 tokens for Hard cases (see Figure~\ref{fig:qvq-analysis}). Incorrect answers are typically even longer, suggesting that excessive reasoning does not guarantee correctness.
\begin{table*}[ht!]
  \centering
  \small
  \setlength{\tabcolsep}{5pt}
  \renewcommand{\arraystretch}{1.15}
  \caption{Quantitative results on the MMMU and MathVista datasets. All results are averaged over 3 random seeds; 95\% confidence intervals are computed via bootstrap resampling.}
  \label{tab:quantitative_results}
  \resizebox{\textwidth}{!}{
    \begin{tabular}{lcccccccccccc}
      \toprule
      \textbf{Dataset} & \makecell{Qwen2.5-\\VL-3B} & \makecell{Qwen2.5-\\VL-7B} & \makecell{Qwen2.5-\\VL-32B} & \makecell{Qwen2.5-\\VL-72B} & \makecell{Gemini-2\\Flash} & GPT-4o & \makecell{Claude3.5\\Sonnet} & \makecell{Qwen2-\\VL-72B} & QVQ & \makecell{Ours\\3B} & \makecell{Ours\\11B} & \makecell{Ours\\7B} \\
      \midrule
      MMMU      & 53.1 & 58.6 & 70.0 & 70.2 & 70.7 & 70.3 & 70.4 & 64.5 & 60.9 & 60.2 & 66.7 & 68.9 \\
      MathVista & 62.3 & 68.2 & 74.7 & 74.8 & 73.1 & 63.8 & 65.4 & 70.5 & 65.4 & 67.1 & 72.0 & 74.2 \\
      \bottomrule
    \end{tabular}
  }
\vspace{-5mm}
\end{table*}
Notably, QVQ exhibits adaptive reasoning: for more difficult questions, it generates longer and more detailed reasoning chains, reflecting increased effort to address complexity. However, our analysis indicates that once token usage exceeds approximately 2,000 tokens, further reasoning does not improve accuracy and may even reduce it. We set a 4,000-token cutoff to avoid excessive computation and latency, as well as to align with practical deployment constraints. Upon manually examining QVQ's incorrect cases, we find that the reasoning steps themselves are often logically sound. However, errors frequently arise during visual readout operations—when revisiting the image, they sometimes produces statements that do not align with the visual content. These failures are commonly due to mistakes in reading numbers, misidentifying details, or other perceptual inaccuracies. Moreover, some hard problems require intensive computation, which consumes a large number of tokens and increases the likelihood of errors.

\subsection{Agent-based Diagnostic and Intervention}

To overcome the limitations of purely LLM-driven visual reasoning, we propose an agent-based architecture that seamlessly combines LLM reasoning with lightweight visual modules. This modular design enables precise analysis and iterative refinement of reasoning chains, allowing us to pinpoint whether failures stem from perceptual errors or reasoning weaknesses. We evaluate our approach across a suite of multimodal tasks, leveraging specialized tools such as \textit{OCR}, \textit{Image Captioning}, \textit{Image Question and answering} and a \textit{Python interpreter}, with \texttt{Qwen2.5-VL-3b,7B}~~\cite{bai2025qwen25vl} serving as the backbone. Our agent’s reasoning mode is denoted as \texttt{qwq}~~\cite{yang2024qwen2}.
Based on the hypothesis that visual grounding errors are a major source of failure, we experiment with agent-based systems using three backbone sizes—3B, 7B, and 11B, with the latter matching LLaVA-CoT.

Our results yield several key insights:
\begin{enumerate*}
  \item \textit{Strong Performance Without Large-Scale Models:} Our agent-based system (“Ours 7B”) achieves 68.9\% on MMMU and 74.2\% on MathVista, rivaling top-tier models such as Qwen2.5-VL-72B, Gemini-2 Flash, and GPT-4o, despite using a much smaller backbone. This demonstrates that modularizing perception and reasoning can yield substantial gains without increasing model size. Notably, the 7B backbone consistently outperforms the larger 11B variant, highlighting the effectiveness of our modular approach.
  \item \textit{Dedicated Visual Tools Enhance Reasoning:} On MathVista, our system matches the performance of much larger models, underscoring that perceptual grounding (e.g., accurate text and layout extraction) is a key bottleneck. Specialized tools such as OCR are essential for these tasks.
  \item \textit{Task-Specific Gains:} On MMMU, our system outperforms the base Qwen2.5-VL-7B by 10.3 points; on MathVista, where perceptual accuracy is critical, the improvement is even greater (+6.0 points). This supports the view that many visual reasoning failures stem from perceptual errors, which modular pipelines can address. Unlike monolithic VL models, our agent architecture enables multi-step reasoning, such as re-querying OCR or cross-checking visual entities with logical constraints, providing greater flexibility and effectiveness without increasing model size.
\end{enumerate*}
We performed analysis on 100 incorrect responses from both baseline and
our models, categorizing errors as OCR, spatial, math.
Baseline errors: OCR (38\%), spatial (22\%), math (19\%).
With our agent, these dropped to OCR (19\%), spatial (15\%), math (13\%).
\subsection{Ablation Study}
\begin{table}[ht]
\centering
\small
\setlength{\tabcolsep}{4pt}
\renewcommand{\arraystretch}{1.1}
\caption{Ablation study of our agent-based system (7B backbone) on MMMU and MathVista. Each column disables a specific module.}
\label{tab:ablation}
\resizebox{\linewidth}{!}{
\begin{tabular}{lcccccc}
  \toprule
  \textbf{Dataset} & \textbf{Full} & \textbf{- OCR} & \textbf{- Python} & \textbf{- Caption} & \textbf{- QA} & \textbf{- Backtrace} \\
  \midrule
  MMMU      & 68.9 & 62.1 & 65.4 & 66.2 & 67.0 & 60.8 \\
  MathVista & 74.2 & 66.7 & 70.3 & 71.1 & 72.0 & 69.5 \\
  \bottomrule
\end{tabular}
}
\vspace{-5mm}
\end{table}

\noindent
Table~\ref{tab:ablation} summarizes the effect of removing each module. Disabling OCR causes the largest drop, especially on MathVista, confirming its critical role. The Python interpreter and captioning modules also yield notable gains, while the QA tool has a smaller effect. Removing backtracing significantly reduces performance, underscoring its importance for error correction for iterative reasoning.

\section{Conclusion and Limitations}
Our diagnostic framework demonstrates that targeting common failure modes enables strong performance even with smaller backbones. Looking forward, \textbf{future visual reasoning models should focus on integrating a broader set of specialized tools for analyzing visual content}. Beyond simply calling external tools, models should natively incorporate these capabilities. This direction will help models better adapt to diverse and complex real-world scenarios.
Our analysis is limited to math-centric visual reasoning, and findings may not generalize to other domains such as document understanding or natural scene understanding.

\newpage
\bibliography{main}

\begin{thebibliography}{22}
\providecommand{\natexlab}[1]{#1}

\bibitem[{hei(2025)}]{heima2025efficient}
 2025.
\newblock \href {https://arxiv.org/abs/2501.19201} {Efficient reasoning with hidden thinking}.
\newblock \emph{Preprint}, arXiv:2501.19201.

\bibitem[{Author(s)(2025)}]{rethinking2025}
Author(s). 2025.
\newblock \href {https://medium.com/@joycebirkins/latest-advances-in-llm-reasoning-4-papers-on-evaluation-reflection-hallucination-and-grm-8d3a13e01045} {Rethinking reflection in pre-training}.
\newblock \emph{Journal Name}.

\bibitem[{Bai et~al.(2025)Bai, Chen, Liu, Wang, Ge, Song, Dang, Wang, Wang, Tang, Zhong, Zhu, Yang, Li, Wan, Wang, Ding, Fu, Xu, Ye, Zhang, Xie, Cheng, Zhang, Yang, Xu, and Lin}]{bai2025qwen25vl}
Shuai Bai, Keqin Chen, Xuejing Liu, Jialin Wang, Wenbin Ge, Sibo Song, Kai Dang, Peng Wang, Shijie Wang, Jun Tang, Humen Zhong, Yuanzhi Zhu, Mingkun Yang, Zhaohai Li, Jianqiang Wan, Pengfei Wang, Wei Ding, Zheren Fu, Yiheng Xu, and 8 others. 2025.
\newblock \href {https://arxiv.org/abs/2502.13923} {Qwen2.5-vl technical report}.
\newblock \emph{arXiv preprint arXiv:2502.13923}.

\bibitem[{Fu et~al.(2025)Fu, Liu, Yang, Corring, Lu, Yang, Roth, Florencio, and Zhang}]{fu2025refocus}
Xingyu Fu, Minqian Liu, Zhengyuan Yang, John Corring, Yijuan Lu, Jianwei Yang, Dan Roth, Dinei Florencio, and Cha Zhang. 2025.
\newblock Refocus: Visual editing as a chain of thought for structured image understanding.
\newblock \emph{arXiv preprint arXiv:2501.05452}.

\bibitem[{Kumar et~al.(2025)Kumar, Kim, Nathani, and Roy}]{improving2025}
Adarsh Kumar, Hwiyoon Kim, Jawahar~Sai Nathani, and Neil Roy. 2025.
\newblock \href {https://arxiv.org/abs/2505.09031} {Improving the reliability of llms: Combining cot, rag, self-consistency, and self-verification}.
\newblock \emph{arXiv preprint arXiv:2505.09031}.

\bibitem[{Kumar et~al.(2024)Kumar, Gadhia, Ganu, and Nambi}]{kumar2024mmctagent}
Somnath Kumar, Yash Gadhia, Tanuja Ganu, and Akshay Nambi. 2024.
\newblock Mmctagent: Multi-modal critical thinking agent framework for complex visual reasoning.
\newblock \emph{arXiv preprint arXiv:2405.18358}.

\bibitem[{Liu et~al.(2025{\natexlab{a}})Liu, Chen, Ding, Xu, Wu, and Wang}]{attention2025}
Qiang Liu, Xinlong Chen, Yue Ding, Shizhen Xu, Shu Wu, and Liang Wang. 2025{\natexlab{a}}.
\newblock \href {https://arxiv.org/abs/2501.09997} {Attention-guided self-reflection for zero-shot hallucination detection in large language models}.
\newblock \emph{arXiv preprint arXiv:2501.09997}.

\bibitem[{Liu et~al.(2025{\natexlab{b}})Liu, Qu, Zhong, Peng, Liu, Yu, and Jia}]{liu2025visionreasoner}
Yuqi Liu, Tianyuan Qu, Zhisheng Zhong, Bohao Peng, Shu Liu, Bei Yu, and Jiaya Jia. 2025{\natexlab{b}}.
\newblock Visionreasoner: Unified visual perception and reasoning via reinforcement learning.
\newblock \emph{arXiv preprint arXiv:2505.12081}.

\bibitem[{Lu et~al.(2025)Lu, Liu, Xu, Nan, Yu, Chen, and Wang}]{auditing2025}
Haolang Lu, Yilian Liu, Jingxin Xu, Guoshun Nan, Yuanlong Yu, Zhican Chen, and Kun Wang. 2025.
\newblock \href {https://arxiv.org/abs/2505.13143} {Auditing meta-cognitive hallucinations in reasoning large language models}.
\newblock \emph{arXiv preprint arXiv:2505.13143}.

\bibitem[{Lu et~al.(2024)Lu, Bansal, Xia, Liu, Li, Hajishirzi, Cheng, Chang, Galley, and Gao}]{lu2024mathvista}
Pan Lu, Hritik Bansal, Tony Xia, Jiacheng Liu, Chunyuan Li, Hannaneh Hajishirzi, Hao Cheng, Kai-Wei Chang, Michel Galley, and Jianfeng Gao. 2024.
\newblock \href {https://arxiv.org/abs/2310.02255} {Mathvista: Evaluating mathematical reasoning of foundation models in visual contexts}.
\newblock In \emph{Proceedings of the International Conference on Learning Representations (ICLR)}.

\bibitem[{Shen et~al.(2025)Shen, Wang, Shi, Wang, Zhao, and Gu}]{shen2025efficient}
Xuan Shen, Yizhou Wang, Xiangxi Shi, Yanzhi Wang, Pu~Zhao, and Jiuxiang Gu. 2025.
\newblock \href {https://arxiv.org/abs/2501.19201} {Efficient reasoning with hidden thinking}.
\newblock \emph{arXiv preprint arXiv:2501.19201}.

\bibitem[{Shi et~al.(2024)Shi, Jiang, Qiu, and Yang}]{shi2024agentre}
Yuchen Shi, Guochao Jiang, Tian Qiu, and Deqing Yang. 2024.
\newblock Agentre: An agent-based framework for navigating complex information landscapes in relation extraction.
\newblock \emph{arXiv preprint arXiv:2409.01854}.

\bibitem[{Song et~al.(2025)Song, Wu, Wang, Liu, Su, and Zheng}]{progco2025}
Xiaoshuai Song, Yanan Wu, Weixun Wang, Jiaheng Liu, Wenbo Su, and Bo~Zheng. 2025.
\newblock \href {https://arxiv.org/pdf/2501.01264} {Progco: Program helps self-correction of large language models}.
\newblock \emph{arXiv preprint arXiv:2501.01264}.

\bibitem[{Team(2024)}]{qvq-72b-preview}
Qwen Team. 2024.
\newblock \href {https://arxiv.org/abs/2409.12191} {Qvq: To see the world with wisdom}.
\newblock \emph{Preprint}, arXiv:2409.12191.

\bibitem[{Thawakar et~al.(2025)Thawakar, Dissanayake, More, Thawkar, Heakl, Ahsan, Li, Zumri, Lahoud, Anwer, Cholakkal, Laptev, Shah, Khan, and Khan}]{thawakar2025llamavo1}
Omkar Thawakar, Dinura Dissanayake, Ketan More, Ritesh Thawkar, Ahmed Heakl, Noor Ahsan, Yuhao Li, Mohammed Zumri, Jean Lahoud, Rao~Muhammad Anwer, Hisham Cholakkal, Ivan Laptev, Mubarak Shah, Fahad~Shahbaz Khan, and Salman Khan. 2025.
\newblock \href {https://arxiv.org/abs/2501.06186} {Llamav-o1: Rethinking step-by-step visual reasoning in llms}.
\newblock \emph{Preprint}, arXiv:2501.06186.

\bibitem[{Xu et~al.(2024)Xu, Jin, Li, Song, Sun, and Yuan}]{xu2024llavacot}
Guowei Xu, Peng Jin, Hao Li, Yibing Song, Lichao Sun, and Li~Yuan. 2024.
\newblock \href {https://arxiv.org/abs/2411.10440} {Llava-cot: Let vision language models reason step-by-step}.
\newblock \emph{Preprint}, arXiv:2411.10440.

\bibitem[{Yang et~al.(2024)Yang, Yang, Zhang, Hui, Zheng, Yu, Li, Liu, Huang, Wei et~al.}]{yang2024qwen2}
An~Yang, Baosong Yang, Beichen Zhang, Binyuan Hui, Bo~Zheng, Bowen Yu, Chengyuan Li, Dayiheng Liu, Fei Huang, Haoran Wei, and 1 others. 2024.
\newblock Qwen2.5 technical report.
\newblock \emph{arXiv preprint arXiv:2412.15115}.

\bibitem[{Yang et~al.(2023)Yang, Li, Li, Zhao, Tan, Du, Yu, Chang, Wu, and Bansal}]{yang2023mmreact}
Diyi Yang, Junnan Li, Xiangru Li, Wayne Zhao, Ming Tan, Jing Du, Zhou Yu, Kai-Wei Chang, Zichao Wu, and Mohit Bansal. 2023.
\newblock Mm-react: Prompting multi-modal chain-of-thought reasoning in language-image models.
\newblock \emph{arXiv preprint arXiv:2303.11381}.

\bibitem[{Yang et~al.(2025)Yang, Luo, Han, and Hovy}]{yang2025magicvqa}
Shuo Yang, Siwen Luo, Soyeon~Caren Han, and Eduard Hovy. 2025.
\newblock Magic-vqa: Multimodal and grounded inference with commonsense knowledge for visual question answering.
\newblock \emph{arXiv preprint arXiv:2503.18491}.

\bibitem[{Yao et~al.(2024)Yao, Huang, Wu, Zhang, Wang, Liu, Wang, Song, Feng, Shen, and Tao}]{yao2024mulberry}
Huanjin Yao, Jiaxing Huang, Wenhao Wu, Jingyi Zhang, Yibo Wang, Shunyu Liu, Yingjie Wang, Yuxin Song, Haocheng Feng, Li~Shen, and Dacheng Tao. 2024.
\newblock \href {https://arxiv.org/abs/2412.18319} {Mulberry: Empowering mllm with o1-like reasoning and reflection via collective monte carlo tree search}.
\newblock \emph{Preprint}, arXiv:2412.18319.

\bibitem[{Yue et~al.(2023)Yue, Ni, Zhang, Zheng, Liu, Zhang, Stevens, Jiang, Ren, Sun, Wei, Yu, Yuan, Sun, Yin, Zheng, Yang, Liu, Huang, Sun, Su, and Chen}]{yue2023mmmu}
Xiang Yue, Yuansheng Ni, Kai Zhang, Tianyu Zheng, Ruoqi Liu, Ge~Zhang, Samuel Stevens, Dongfu Jiang, Weiming Ren, Yuxuan Sun, Cong Wei, Botao Yu, Ruibin Yuan, Renliang Sun, Ming Yin, Boyuan Zheng, Zhenzhu Yang, Yibo Liu, Wenhao Huang, and 3 others. 2023.
\newblock \href {https://arxiv.org/abs/2311.16502} {Mmmu: A massive multi-discipline multimodal understanding and reasoning benchmark for expert agi}.
\newblock \emph{Preprint}, arXiv:2311.16502.

\bibitem[{Zhang et~al.(2025)Zhang, Zhang, Ju, Liu, Mao, Sun, Wu, Gao, Cai, Qin, Liang, Wang, Duan, Cao, Xu, and Tang}]{zhang2025embodiedvsr}
Yi~Zhang, Qiang Zhang, Xiaozhu Ju, Zhaoyang Liu, Jilei Mao, Jingkai Sun, Jintao Wu, Shixiong Gao, Shihan Cai, Zhiyuan Qin, Linkai Liang, Jiaxu Wang, Yiqun Duan, Jiahang Cao, Renjing Xu, and Jian Tang. 2025.
\newblock Embodiedvsr: Dynamic scene graph-guided chain-of-thought reasoning for visual spatial tasks.
\newblock \emph{arXiv preprint arXiv:2503.11089}.

\end{thebibliography}
\end{document}